\documentclass[11pt,a4paper]{article}
\usepackage[hyperref]{acl2021}
\usepackage{times}
\usepackage{latexsym}

\usepackage{amsmath,graphicx,amsfonts}
\usepackage[T1]{fontenc}
\usepackage{titlesec,comment,multirow}
\usepackage{multirow}
\aclfinalcopy
\title{The IWSLT 2021 BUT Speech Translation Systems}
 \author{Hari Krishna Vydana, Martin Karafiát, Luk\'{a}\v{s} Burget, ``Honza'' \v{C}ernock\'{y} \\
 Brno University of Technology, Faculty of Information Technology, Speech@FIT, Czechia
 \\
 \texttt{harivydana@gmail.com}}
\date{}
\begin{document}
%\ninept
\maketitle
\begin{abstract}
The paper describes BUT's English to German offline speech translation(ST) systems developed for IWSLT2021. They are based on jointly trained Automatic Speech Recognition-Machine Translation models. Their  performances is evaluated on MustC-Common test set.
In this work, we study their  efficiency from the perspective of having a large amount of separate ASR training data and MT training data, and a smaller amount of speech-translation training data. Large amounts of ASR and MT training data are utilized for pre-training the ASR and MT models. Speech-translation data is used to jointly optimize ASR-MT models by defining an end-to-end differentiable path from speech to translations. For this purpose, we use the internal continuous representations from the ASR-decoder as the input to MT module. We show that speech translation can be further improved by training the ASR-decoder jointly with the MT-module using large amount of text-only MT training data. We also show significant improvements by training an ASR module capable of generating punctuated text, rather than leaving the punctuation task to the MT module. 

%We also explore an adaptation scheme to increase tight-coupling between ASR and MT model and this has improved the performance.
%These systems are based on our previous work in ~\cite{vydana2020jointly}.
\end{abstract}

\section{Introduction}
\label{sec:intro}

Speech Translation (ST) systems are intended to generate text in target language from the audio in source language. The conventional ST systems are cascade ones, including (in the most popular form)  three blocks i.e., an ASR, punctuation/segmentation module and an MT model~\cite{KIT_IWSLT2019, KIT_IWSLT2020, IWSLT2019_eval_campaign, IWSLT2020_eval_campaign}. Both Automatic Speech Recognition system~(ASR) and Machine Translation~(MT) models are independently trained, and the MT model processes the ASR output text (ASR hypotheses) to generate translations. In a cascade system, the advancements in ASR and MT can be directly extended to ST. These models can also leverage on the availability of large ASR and MT data-sets, and some of the state-of-the art ST systems are still cascade ones.  

Recently,  End-to-End ST systems have become widely popular. An End-to-End ST can directly generate text in target language from the audio in source language. These models are simpler in structure and they are more suitable for operating in streaming fashion.
Most End-to-End speech translation systems are variants of encoder-decoder architecture with attention models~\cite{neural_MT,di2019adapting,neural_st_toolkit}.  This category includes the popular Transformer models, which have been adapted for training End-to-End ST in~\cite{di2019adapting}. In ~\cite{Espnet_ST_toolkit}, a better performance of ST was achieved by initializing the encoder and decoder modules from pre-trainied ASR and MT systems, respectively. 
Very-deep transformer models have been trained with stochastic depth for training End-to-End ST models in~\cite{Transformer_stocastic_depth}. The use of relative positional embeddings has also improved the performance of transformer~\cite{relative_position_embedings}. 

One major drawback or end-to-end ST is the data availability, i.e.,  paired speech-to-translation data is scarce compared to ASR or MT data. Data augmentations and use of synthetic data have been explored in~\cite{Spech_Aug_speech_translation,aachen_IWSLT_2010} to mitigate the issue. Unlike End-to-End ST systems, the data for training cascade systems is easily available and less costly.

A brief survey of existing approaches and their principal limitations are discussed in~\cite{Apple_review_paper_tight_coupling}. Despite multiple advantages, the cascade systems suffer from a major drawback:  propagating erroneous early decisions into MT models, which then cause degradation in the translation performance. To mitigate this degradation, rather than passing a single ASR output sequence to MT model, other forms such as lattices, n-best hypotheses and continuous representations have been explored in ~\cite{passing_continous_vectors,lattice_transformer,sperber2019attention,vydana2020jointly,triple_supervison}. 

In this work, we use our jointly trained Automatic Speech Recognition-Machine Translation~(Joint-ASR-MT) model previously described in~\cite{vydana2020jointly}. Joint-ASR-MT model is a cascade system, but it has a differentiable path between ASR and MT modules. To create such differentible path, the continuous hidden representations (corresponding to each output token) from the ASR decoder are passed to the MT-Model. The hidden continuous tokens corresponding to each output token are the attention-weighted value vectors in the last layer of the transformer decoder. We refer to these continuous representations as``context vectors'' as proposed in~\cite{sperber2019attention}. 

Existing large separate ASR training data and MT training data can be used to pre-train these modules; then, the pre-trained modules are jointly optimized using a small amount of speech translation data. The joint optimization mitigates the degradation in performance due to erroneous early decisions. 

In this paper, we generate German translation from English speech, and we focus on two main contributions: (1) We train different MT models that can translate normalized text or punctuated text. It is known that MT-models translating  punctuated text provide superior performance, therefore, we propose to train an ASR system that can generate the punctuated text. We confirm that such ASR system provides superior performance in ASR-MT pipeline. 
(2) 
%Speech-translation data is used to jointly optimize ASR-MT models by having an end-to-end differentiable path form speech to translations. For this purpose, 
We use the internal continuous representations from the ASR-decoder as the input to MT module. In section~\ref{tight_coupling}, we show that speech translation can be further improved by adapting ASR-decoder to the MT module. This is achieved by training the ASR-decoder jointly with the MT-module using a large amount of text-only MT training data.

%The data used for training the models is described in section~\ref{sec:data}. The training and performance of ASR, MT models are described in sections ~\ref{section_Transformer_ASR}, ~\ref{section_Transformer_machiene_translation} respectively. Joint training of ASR and MT systems are described in section~\ref{section_joint_ASR_MT}. The conclusions from the work are presented in section~\ref{section_conclusion}.

\section{Datasets and Pre-processing}
\label{sec:data}
The Datasets used for training various models are described in Table.~\ref{training_datasets}. ASR-Train-set and MT-Train-set are used for pre-training ASR and MT models respectively. The pre-trained models are fine-tuned using ASR-MT-Train-set. All models are evaluated using MustC-Common test set.

\begin{table}[!ht]
\renewcommand{\arraystretch}{1.5}
\setlength{\tabcolsep}{1.5pt}
\scriptsize
\begin{center}
\caption{Data used for training various models.}
\label{training_datasets}
\begin{tabular}{|l|l|c|c|c|c|}
\hline
~&Corpora & \#Sentences&Audio&\shortstack{Source\\text}&\shortstack{Target\\Text}\\
\hline
\multirow{2}{*}{\shortstack{MT\\-Train-set}}&ParaCrawl v3&31M&-&\checkmark&\checkmark\\ 
~&OpenSubtitles 2018&12M&-&\checkmark&\checkmark\\
~&Rapid 2019 &1.5M&-&\checkmark&\checkmark\\
~&Europarl v9 &1.81M&-&\checkmark&\checkmark\\
~&News Commentary &365K&-&\checkmark&\checkmark\\
~&Common Crawl &2.4M&-&\checkmark&\checkmark\\
~&Wikititles &1.3M&-&\checkmark&\checkmark\\
~&WIT3 &196K&-&\checkmark&\checkmark\\
~&TED Talks&220K&-&\checkmark&\checkmark\\
\hline
\multirow{2}{*}{\shortstack{ASR-MT\\-Train-set}}&Europarl-ST &32K&\checkmark&\checkmark&\checkmark\\
~&Must-C V2&230K&\checkmark&\checkmark&\checkmark\\
~&IWSLT2018 &171K&\checkmark&\checkmark&\checkmark\\
\hline
\multirow{2}{*}{\shortstack{ASR\\-Train-set}}&Tedlium3 &264K&\checkmark&\checkmark&-\\
~&Librispeech&281K&\checkmark&\checkmark&-\\
\hline
\end{tabular}
\end{center}
\end{table}

\subsection{Pre-processing and Feature Extraction}
\label{preprocessing}
From audio data, 80-Dimensional Mel-Filter bank energies along with pitch features are extracted. The Moses toolkit is used for text tokenization and other standard text pre-processing. 
The umlauts from the German text are replaced by the special tokens. All the non ASCII characters are removed from the text data. The repetitions of the same sentences are removed from the corpora. We cleaned up the MT training data by identifying and manually removing the sentences where successive words were erroneously concatenated in to very long erroneous words. Sentence-piece models~\cite{sentencepiece_model} are used for training BPE-tokenizers. 40M lines of text are used for training each BPE-tokenizer and all the tokenizers have a vocabulary of 20K units. Three separate tokenizers are trained using normalized English text, punctuated English text and punctuated German text. The output of MT module is always punctuated text, while input to MT (as well as ASR output) can be either normalized or punctuated text~(see norm-MT and Punc-MT in sections~\ref{section_Transformer_machiene_translation}).   

%---------------------------------------------------------------------------
\subsection{Pruning Noisy ASR corpus}
\label{pruning}
Some of the utterances in ASR-MT-Train-set~(MustC, IWSLT and Europarl) sets are erroneous due to the shift in alignments between audio and text. Training an End-to-End ASR on this data directly did not lead to convergence. To remove erroneous transcripts, a hybrid TDNN-LFMMI ASR system based on KALDI~\cite{Kaldi_toolkit, povey2016_lfmmi}  was trained and this ASR system was used to decode the ASR-MT-train set. The Word Error Rate (WER) for each sentence is computed and the sentences with more than 50\% WER are deleted from the ASR-MT-Train-set~\cite{samsung_2019}. Even with this cleaning, training the ASR systems only on ASR-MT-Train-set did not lead to convergence. Pre-training the ASR models on ASR-Train-set turned out to be crucial for convergence as described in section~\ref{section_Transformer_ASR}.

%-----------------------------------------------
\section{Automatic Speech Recognition~(ASR)}
\label{section_Transformer_ASR}
ASR systems trained in this work are built on Transformer~ASR models~\cite{Speech-transformer,ESPNET_rnn_tranformer,vydana2020jointly,Transformer}. The ASR models have 12 encoder and 6 decoder layers with 4096 feed-forward units and 1024 attention dimension with 16 heads. Models are initially trained with ASR-Train-set and are later fine-tuned with ASR-MT-Train-set. A thresholding mechanism is used for pruning away the noisy end-of-sequence~(EOS) tokens from beam search~\cite{selftraining_EOS_threshold_FAIR}. Models are trained with 30K warm-up updates and a check-point is saved after every 8K updates. The training is stopped with an early stopping criterion. 8-best check-points are averaged and the averaged weights are used for decoding the hypothesis. Vectorized beam search~\cite{vectorized_beam_search} was used for decoding the ASR hypotheses with a beam size of 10.  Further in this paper, ASR models described in this section are referred to as Ext.ASR models (Externally trained ASR models).

Two different ASR systems were trained for generating normalized text (Norm-ASR) and punctuated text (Punc-ASR), and  their performances are reported in  Table~\ref{ASR_results}. It can be observed that the WER of Punc-ASR appears to be higher than Norm-ASR. Punc-ASR is a obviously more difficult task than Norm-ASR --- the  punctuation tokens are considered as extra words and each error in those words contributes to the WER. 
\begin{table}[!ht]
\caption{Performance of trained ASR systems reported on MustC-Common set. For Punc-ASR, the errors in punctuation tokens are considered, which makes it a more difficult task.}
\label{ASR_results}
\centering
\begin{tabular}{lc}
\hline
Model&WER\\
\hline
Norm-ASR&18.20\\
\hfill+LM& 17.35\\
\hline
\hline
Punc-ASR & 21.20\\
\hline
\end{tabular}
\end{table}

\paragraph{ASR-LM:}
\label{ASR_LM}
A Transformer language model was trained on English text~\cite{irie2019language}. The model has 6 layers, with 4096 feed-forward units and 1024 attention dimension with 8 heads. The model is initially pre-trained on Librispeech LM corpus and it is later fine-tuned on English text from MT-train-set and ASR-MT-train-set. An improvement in the performance is observed by shallow fusion of the ASR and language model (ASR-LM). Performances of these language models are presented in column 2 of Table.~\ref{Joint_ASR_MT_model_nbest}.

\section{Machine Translation Systems(MT)}
\label{section_Transformer_machiene_translation}
Transformer models~\cite{Transformer} are also at the core of MT-systems. They have 6-encoder and 6-decoder layers with 4096 feed-forward units and 1024 attention dimensions and have 16 heads. The models are optimized with 30K warm-up updates and a check-point is saved every 8k updates. Training is stopped using an early stopping criterion.  8-best check-points are averaged and the averaged weights are used for decoding the hypotheses. The noisy EOS tokens are pruned out using ~\cite{selftraining_EOS_threshold_FAIR}. Vectorized beam~\cite{vectorized_beam_search} search has been used for decoding the hypotheses with a beam size of 8. A large variance in the performance is observed w.r.t the decoding hyper-parameters such as maximum target sequence length and length-bonus. The maximum length of the target sequence is computed by multiplying the input sequence length with length-ratio: 1.2 was found as optimal on the development set.
To control the length of the output sequence, the log-likelihood scores of the hypotheses are penalized by additive token insertion penalties. The optimal value for this penalty is tuned as a hyper-parameter on the development set. The hypothesis text is de-tokenized and BLEU score is evaluated using Moses Toolkit. All the BLEU scores reported in this paper are computed using the de-tokenized, punctuated German text using {\tt multi-bleu-detok.perl}. The performances of the MT systems are reported in Table.~\ref{MT_results}. All BLEU scores reported in this paper are computed using punctuated text as reference.

\begin{table}[!ht]
\centering
\caption{Performances of the MT systems reported on MustC-Common set.}
\label{MT_results}
\begin{tabular}{lc}
\hline
Model&BLEU\\
\hline
Norm-MT &~\\
\hfill +pretrain & 27.18\\
\hfill +finetune & 27.98\\
\hfill +MT-LM & 28.12\\
\hline
Punc-MT&~\\
\hfill +pretrain & 31.02\\
\hfill +finetune & 35.00\\
\hfill +MT-LM & 35.04\\
\hline
%Punc-MT1 & 30.35\\
%\hfill + Norm-eval & 22.44\\
\hline
\end{tabular}
\end{table}

In Table~\ref{MT_results}, Norm-MT, Punc-MT are MT models trained to predict punctuated German text. Norm-MT uses the normalized English text as input while the Punc-MT uses the punctuated English text. Punc-MT model has performed better than Norm-MT. From  Table~\ref{MT_results}, it can be observed that the punctuation tokens in the text are adding additional information for training the MT model. Fine-tuning the Punc-MT on in-domain text has improved the performance significantly. Further in this paper, MT models described in this section are referred to as Ext.MT models (Externally trained MT models).

\paragraph{MT-LM:}
A transformer language model has been trained on German text from MT-Train-set, ASR-MT-train-set. This LM is also used while decoding with the MT model~\cite{irie2019language}. The architecture of the model is same as ASR-LM mentioned in section~\ref{ASR_LM}. A shallow fusion between the MT-model and the MT-LM Language model is performed.  As shown in Table~\ref{MT_results} and column 2 of Table~\ref{Joint_ASR_MT_model_nbest}, the additional language model~(MT-LM) did not improve the performance significantly.

%The weights from Norm-MT are initialized and are finetuned with punctuated text and the performance is reported as Punc-MT1. When the Punc-MT1 model is evaluated using normalized text the performance is presented as Norm-eval in Table~\ref{MT_results}. Form this result, it can be observed that the model trained on punctuated text has shown a large degradation when evaluated on normalized text.

\section{Jointly Trained ASR-MT Systems}
\label{section_joint_ASR_MT}
\begin{figure}[!htbp]
	\begin{flushleft}
	\includegraphics[width=75mm,height=90mm]{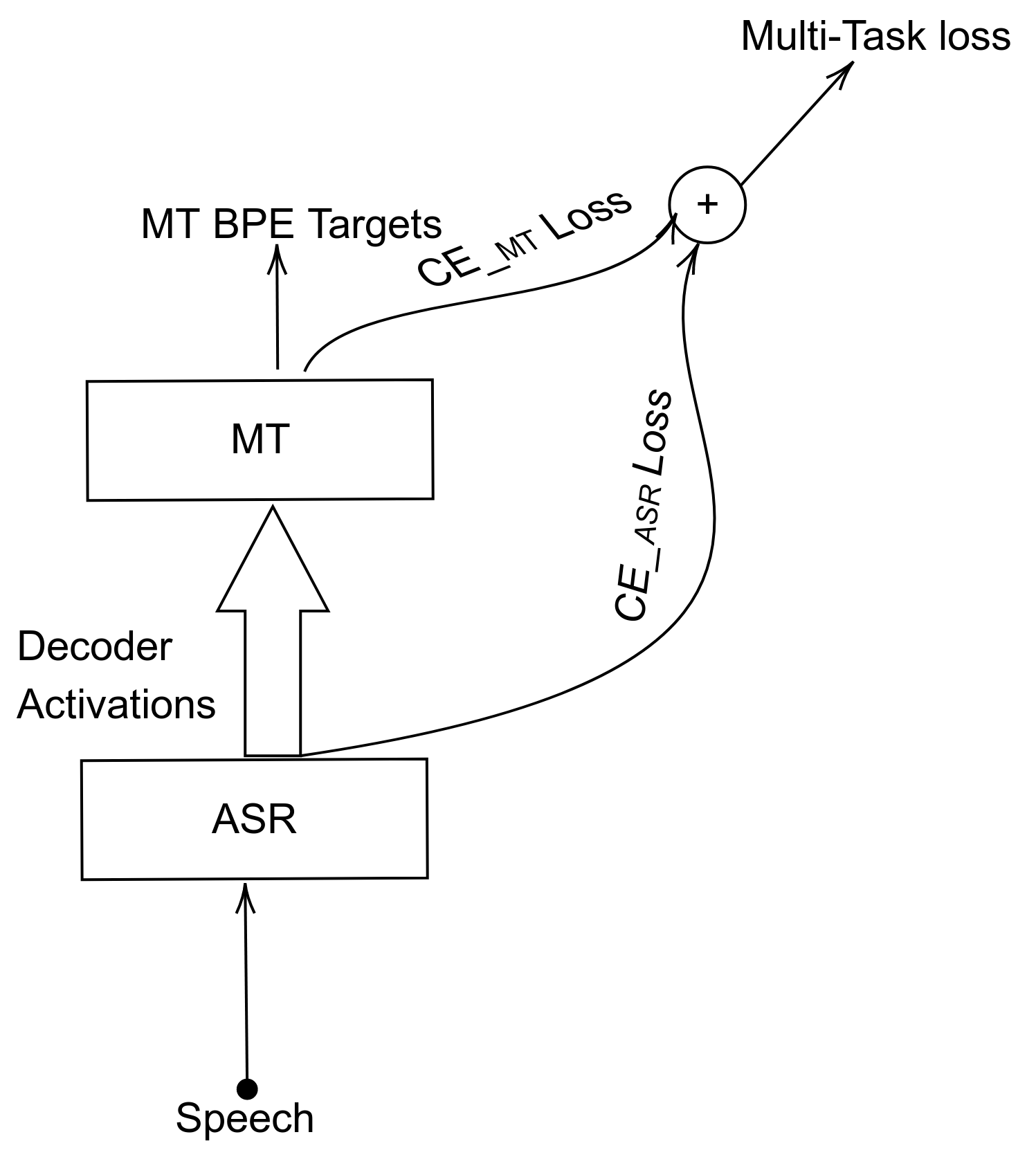}
	\caption{Joint-training of ASR-MT system using multi-task loss.}
	\label{Multi_task_training_with_ASR_auxilary_loss}
	\end{flushleft}
\end{figure}

The model has two modules: ASR and MT; their architecture is same as described in sections~\ref{section_Transformer_ASR} and ~\ref{section_Transformer_machiene_translation} respectively -- see block diagram in Figure~\ref{Multi_task_training_with_ASR_auxilary_loss} and full description of the model in~\cite{vydana2020jointly}.   The context vectors from the final layer of the ASR-decoder are used as inputs to the MT module. Passing context vectors from ASR to MT models while training has also been explored in~\cite{sperber2019attention}. Both the models are jointly optimized using a multi-task cross-entropy (ASR cross-entropy and MT cross-entropy) -- both losses are also  shown in Figure~\ref{Multi_task_training_with_ASR_auxilary_loss}.
During the inference, beam search has been used to obtain the ASR hypotheses, and the corresponding context vectors obtained from the ASR model are used by the MT model for generating translations. The MT model also uses a beam search, and the final ST hypotheses is obtained by a coupled search~\cite{vydana2020jointly} using the joint-likelihood from ASR and MT:   
\begin{align}
\label{coupling_decoders}
y* &=\arg\max_{y}\sum_{z \in \hat{Z}(x)}P(y|z)P(z|x) \nonumber\\
&\equiv \arg\max_{y}\,\,\arg\max_{z \in \hat{Z}(x)}(\log(P(y|z)) \nonumber\\
&{} \qquad+\log(P(z|x))),
\end{align}
where $x$ is the speech  abnd $z$,$y$ are the source and target sequences respectively. $\hat{Z}$ is the n-best source sequence and $y*$ is most likely decoded hypothesis. In this equation, $y*$ is always a discrete sequence, while $z$ is a discrete sequence when we are using Ext.MT and a continuous one when using Joint-MT. Note that similar coupled search was used in~\cite{NMT_backtranslation_loss}, where the back translation likelihoods are used for re-scoring the hypothesis of the MT-system. 
%========================================================================================
\section{Adapting ASR decoder to the MT module} 
\label{tight_coupling}
%We adapt the  ASR-decoder to the MT module. by training the ASR-decoder jointly with the MT-module using large amount of text-only MT training data.  ... HOnza; don't say it twice ... 
Joint-ASR-MT models are jointly optimized by having an end-to-end differentiable path from speech to translations. The internal continuous representations from the ASR-decoder are used as the input to MT module. Speech translation can be further improved by adapting ASR-decoder to the MT module. This is achieved by training the ASR-decoder jointly with the MT-module using large amount of text-only MT training data. The weights for the model are initialized from trained Joint-ASR-MT model. Speech translation data~(ASR-MT-Train-set) is used to fine-tune Joint-ASR-MT model using a multi-task loss. Apart from that, the data from the MT-Train-set is used to jointly train the ASR-decoder and the MT-module of Joint-ASR-MT model. We alternately update the model using multi-task loss described in section~\ref{Joint_ASR_MT_model} and the adaptation loss as described in this section.

\begin{figure}[!h]
	\begin{center}
	\includegraphics[width=75mm,height=35mm]{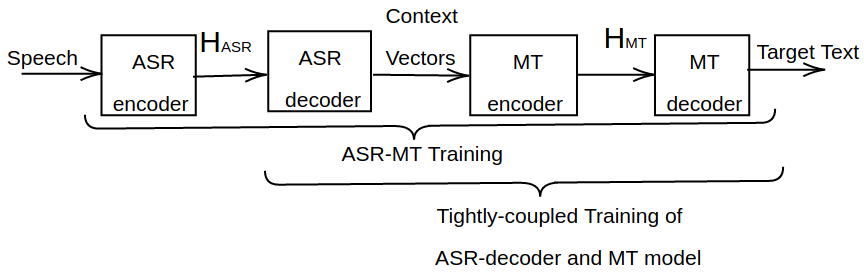}
	\caption{Adaptation of ASR-decoder to the MT-module in the Joint-ASR-MT model.}
	\label{tight_coupling_loss}
	\end{center}
\end{figure}

A block diagram describing this training is presented in Figure~\ref{tight_coupling_loss}.
The input text sequence is given to the ASR-decoder and a sequence of zeros is considered as the encoder output sequence of the ASR model~(i.e.,$H_{ASR}$ in Figure~\ref{tight_coupling_loss}). The context vectors computed from these two sequences are used for training the MT-module. Note that similar method has been adopted in~\cite{samsung_2019} for improving the performance of ASR system using  only text data. This training further improves the performance as will be shown in section~\ref{sec:SpeechTranslationResults}.

%=============================================================================================
\section{Speech Translation Results}
\label{sec:SpeechTranslationResults}
\begin{table*}[!ht]
\renewcommand{\arraystretch}{2}
\setlength{\tabcolsep}{3pt}
\footnotesize%\scriptsize
\caption{Performances of Joint-ASR-MT systems under various ensemble combinations, the results are reported on MustC-Common test set.}
\label{Joint_ASR_MT_model}
\begin{center}
\begin{tabular}{|l|l|cc|cc|cc|cc|}
\hline
~&~&\multicolumn{2}{c}{A}&\multicolumn{2}{|c|}{B}&\multicolumn{2}{|c|}{C}&\multicolumn{2}{|c|}{D}\\
\hline
~&~&\multicolumn{2}{c}{\shortstack{no-pretraining\\+Norm-ASR/MT}}&\multicolumn{2}{|c|}{\shortstack{pre-training\\+Norm-ASR/MT}}&\multicolumn{2}{|c|}{\shortstack{pre-training\\+Punc-ASR/MT}}&\multicolumn{2}{|c|}{\shortstack{pretraining\\+Punc-ASR/MT+\\ tightly-coupled}}\\
%\hline
%\shortstacck{S.No}
&[ASR]$\Rightarrow$[MT]&BLEU&WER&BLEU&WER&BLEU&WER&BLEU&WER\\
\hline
1.&[Ext-ASR]$\Rightarrow$[Ext-MT]&23.20&18.20&23.20&18.20&26.15&21.54&26.15&21.54\\

2.&[Ext-ASR]$\Rightarrow$[Joint-MT]&20.19&-&22.59&- &28.56&-&29.00&-\\

3.&[Ext-ASR]$\Rightarrow$[Joint-MT\,+\,Ext-MT]&24.02&-&24.13&-&29.07&-&29.44&-\\

\hline
4.&[Joint-ASR]$\Rightarrow$[Ext-MT]&23.86&16.14&23.86&13.01&29.70&15.71&30.24&15.63\\ 

5.&[Joint-ASR]$\Rightarrow$[Joint-MT]&20.75&-&23.97&-&31.23&-&32.68&-\\

6.&[Joint-ASR]$\Rightarrow$[Joint-MT\,+\,Ext-MT]&24.65&-&25.95&-&32.51&-&33.68&-\\
\hline
7.&[Ext-ASR\,+\,Joint-ASR]$\Rightarrow$[Ext-MT]&24.60&14.84&25.00&13.54&29.00&16.46&29.35&16.19\\
8.&[Ext-ASR\,+\,Joint-ASR]$\Rightarrow$[Joint-MT]&20.89&-&23.59&-&30.52&-&31.58&-\\

9.&[Ext-ASR\,+\,Joint-ASR]$\Rightarrow$[Joint-MT\,+\,Ext-MT]&25.11&-&25.65&-&31.86&-&32.67&-\\

\hline
\shortstack{10.}&\shortstack{[Ext-ASR\,+\,Joint-ASR]$\Rightarrow$[Joint-MT\,+\,Ext-MT]\\+ens*}&25.35&14.61&26.14&13.05&32.67&15.71&33.78&15.63\\
\hline
\end{tabular}
\end{center}
\end{table*}

%%%%%%%%%%%%%%%%%%%%%%%%%%%%%%%%%%%%%%%%%%%%%%%%%%%%%%%%%%%%%%%%%%%%%%%%%%%%%%%
\begin{table*}[!ht]
\renewcommand{\arraystretch}{1.5}
\setlength{\tabcolsep}{3pt}
\footnotesize%\scriptsize
\caption{Comparing the performance of Joint-ASR-MT systems while processing n-best hypotheses from the ASR.}
\label{Joint_ASR_MT_model_nbest}
\begin{center}
\begin{tabular}{|l|cc|cc|cc|cc|}
\hline
~&\multicolumn{2}{c}{A}&\multicolumn{2}{|c|}{B}&\multicolumn{2}{|c|}{C}&\multicolumn{2}{|c|}{D}\\
\hline
~&\multicolumn{2}{c}{\shortstack{no-pretraining\\+Norm-ASR/MT}}&\multicolumn{2}{|c|}{\shortstack{pre-training\\+Norm-ASR/MT}}&\multicolumn{2}{|c|}{\shortstack{pre-training\\+Punc-ASR/MT}}&\multicolumn{2}{|c|}{\shortstack{pretraining\\+Punc-ASR/MT+\\ tightly-coupled}}\\

[ASR]$\Rightarrow$[MT]&BLEU&WER&BLEU&WER&BLEU&WER&BLEU&WER\\
\hline
\shortstack{[Ext-ASR\,+\,Joint-ASR]$\Rightarrow$[Joint-MT\,+\,Ext-MT]\\+ens*}&25.35&14.61&26.14&13.05&32.67&15.71&33.78&15.63\\

\multicolumn{1}{|c|}{+ASR-LM}&~&~&26.90&12.80&~&~&~&~\\

\multicolumn{1}{|c|}{+MT-LM}&~&~&27.16&~&~&~&~&~\\

\multicolumn{1}{|c|}{+2-best-input}&-&-&27.24&-&32.69&-&33.82&-\\

\multicolumn{1}{|c|}{+4-best-input}&-&-&27.35&-&32.80&-&33.87&-\\

\multicolumn{1}{|c|}{+6-best-input}&-&-&27.35&-&32.85&-&33.86&-\\

\multicolumn{1}{|c|}{+8-best-input}&-&-&27.46&-&32.94&-&33.77&-\\

\multicolumn{1}{|c|}{+10-best-input}&-&-&27.51&-&32.87&-&33.79&-\\
\hline
\end{tabular}
\end{center}
\end{table*}
%%%%%%%%%%%%%%%%%%%%%%%%%%%%%%%%%%%%%%%%%%%%%%%%%%%%%%%%%%%%%%%%%%%%%%%%%%%%%%%%%
Results for the various configurations of speech translation systems are given in Table~\ref{Joint_ASR_MT_model}. First, we focus on column A, where the Joint-ASR-MT models are trained using ASR-MT-Train-set~(only speech translation data) with a multi-task loss as described in section~\ref{section_joint_ASR_MT}. Note, however, that Ext.ASR and Ext.MT systems are trained on large amounts of data and finetuned to ASR-MT-Train-set as described in~sections~\ref{section_Transformer_ASR} and \ref{section_Transformer_machiene_translation} respectively. For systems in column-A, normalized~(unpunctuated) text is passed from ASR to MT model.
%%%%%%%%%%%%%%%%%%%%%%%%%%%%%%%%%%%%%%%%%%%%%%%%%%%%%%%%%%%%%%%%%%%%%%%%%%%%%%%%%%%%%%%%%%%
%====================================================
Row~1 corresponds to the conventional cascade system, where the Ext.ASR systems generates the n-best hypotheses of discrete token sequences and an Ext.MT uses these token sequences for generating the translations as described in Eq.~\ref{coupling_decoders}. We consider this system achieving BLEU 23.20 as a baseline. 

Usually, transformer-ASR decoder uses the partial output hypothesis and extends it by a new token with every autoregressive decoding step. For the system in row~2, Ext.ASR generates the complete hypothesis and ASR module from Joint-ASR-MT is ``asked'' to extend it by one more token. As a byproduct ``context vectors'' (the continuous representations) are generated for the whole sequence --- these are then passed to the MT-module in joint-ASR-MT model to generate translation. Compared to row~1 of column~A, we see a degradation in performance~(BLEU-20.19). This can be attributed to having only small amount of speech translation training data, which is not sufficient for robustly training the Joint-ASR-MT systems.

For the systems in row~3, Ext.ASR generates the ASR hypotheses which are used by Ext.MT similar to the system described in row~1; the hypotheses from Ext.ASR are used by Joint-MT similarly to the system described in row~2. To generate the translation, the hypotheses form both  models are ensembled as follows:  
For each output token, a weighted average of Log-softmax outputs from the two MT models is computed. This weighted average is used in the beam-search to compute the n-best partial hypotheses. These partial hypotheses are further extended by both the models to generate the Log-softmax outputs for next tokens. We can see that this ensembling system achieves a BLEU score of 24.02 and outperforms the cascaded baseline.

%%%%%%%%%%%%%%%%%%%%%%%%%%%%%%%%%%%%%%%%%%%%%%%%%%%%%%%%%%%%%%%%%
The systems in rows~4-6 are essentially the same as the ones in rows~1-3, respectively, except that now, the ASR module from joint-ASR-MT system is directly used to produce the n-best ASR hypotheses and the corresponding context vectors. Rows~4-6 show the same trend as rows~1-3 with slightly improved performance; these improvements are mainly due to better performing ASR system: 
As described in Section~\ref{pruning}, training ASR systems only on ASR-MT-Train-set~(data from Mustc, IWSLT and Europarl with erroneous transcriptions) did not lead to convergence. However, when the same data is used to train Joint-ASR-MT model for speech translation task, we observe that the ASR module in this model trained well. The reason for that is that the ASR-module is not directly trained on erroneous transcriptions, instead, it is trained to produce transcriptions that lead to good translations. This training can be seen as a form of light supervision which can mitigate the problem with the erroneous transcriptions. At the end, this system trained only on ASR-MT-Train-set achieves better ASR performance (WER 16.14\%) compared to Ext.ASR~(WER 18.20\%), Which is pre-trained on ASR-Train-set (Approx 2000hrs) and fine-tuned on erroneous ASR-MT-Train-set. Similar trend will be observed with the systems in columns B, C and D.

%%%%%%%%%%%%%%%%%%%%%%%%%%%%%%%%%%%%%%%%%%%%%%%%%%%%%%%%%%%%%%%%%
The systems described in rows 7-9 are similar to those from rows 1-3, except that the ASR hypotheses are obtained by  ensembling the Ext.ASR and ASR-module in Joint-ASR-MT model. The ensembling is performed in a similar way as described for the MT-system~(row 2).  
All the ensemble systems in rows~3, 6, and 7-9 are ensembled giving equal weight to both the systems, except for row~10, where the ensemble weights are tuned on the development set. For all these systems, we can see that the ensembling consistently improves the performances.

%%%%%%%%%%%%%%%%%%%%%%%%%%%%%%%%%%%%%%%%%%%%%%%%%%%%%%%%%%%%%%%%%
The systems in column~B are similar to the ones in Column~A, but for the Joint-ASR-MT model, the weights of ASR and MT module are initialized from the Ext.ASR and Ext.MT. Only then, the Joint-ASR-MT model is fine-tuned using ASR-MT-Train-set. Comparing column-A and column-B, we can see that such pre-training has significantly improved the performance.

%When the Joint-ASR-MT models are trained on only speech translation data~(ASR-MT-Train-set) with out any pre-training, Ext.MT has performed better that using Joint-MT, as large additional data is used for training Ext.MT. This can be observed by comparing row~1, 2 of column-A and Rows~4, 5 of column~A. 
%Pre-training the ASR, MT modules of Joint-ASR-MT models have significantly improved the performance of ASR and speech translation systems. This improvement can be observed by comparing column-A and column-B.

%When trained on same amounts of data, the performance of MT-model that uses continuous input~(MT-module in Joint ASR-MT model) has performed better than the model that used discrete tokens as input~(Ext.MT). This can be observed by comparing row~4 and row~5 of column~B. Similar trend can also observed in row4, and row~5 in column~C and D.

We also see that the MT system using continuous representations~(Joint-MT)~(row~5; BLEU 23.97) outperforms the system with the Ext.MT (row~4; BLEU-23.86) and similar trend can be seen in columns~C and D. This is in contrast to the system in column~A where we did not use enough data for training the Joint-ASR-MT model; now, with the pre-training, the joint-ASR-MT model is effectively trained on the same amount of data as the Ext.MT systems.

%%%%%%%%%%%%%%%%%%%%%%%%%%%%%%%%%%%%%%%%%%%%%%%%%%%%%%%%%%%%%%%%%
The systems in column~C are similar to the ones in Column~B, but the ASR and MT modules used here are Punc-ASR~(ASR systems which can generate punctuated text) and Punc-MT~(MT systems which can process punctuated text as input), respectively. We can see that the systems from column-C perform significantly and consistently better than the corresponding ones in column-B. This shows that it is more effective to train an ASR module to generate punctuated text rather than leaving the punctuation task to the MT module. Note that the ASR performances reported in columns C and D is computed including the punctuation symbols, which results in higher WERs. 

%%%%%%%%%%%%%%%%%%%%%%%%%%%%%%%%%%%%%%%%%%%%%%%%%%%%%%%%%%%%%%%%%

Finally, the systems in column~D are the same as the ones in column~C except that we additionally use the ASR decoder adaptation scheme described in section~\ref{tight_coupling}. The consistent improvements observed in column~D as compared to column~C show the effectiveness of this adaptation scheme. They are able to make use of the large amount of text-only MT training data to train also the ASR decoder in order to tighten the coupling between ASR-decoder and MT-module. 
Apart from improving MT-module,  this adaptation has also improved the performance of ASR-decoder on its own. This can be observed by comparing WER's of row 4 in columns C and D. 

%%%%%%%%%%%%%%%%%%%%%%%%%%%%%%%%%%%%%%%%%%%%%%%%%%%%%%%%%%%%%%%%%

The results of passing the n-best hypotheses from ASR to MT models are presented in Table~\ref{Joint_ASR_MT_model_nbest}. Passing the n-best hypothesis from ASR to MT module has better performance, but not significantly. This result is not in line with out previous studies~\cite{vydana2020jointly}, where we have seen significant gains from switching from 1-best to n-best.

\section{Conclusion}
\label{section_conclusion}
In this work, we have explored joint-training of ASR-MT models for speech translation. Initializing these models from pre-trained ASR and MT models has helped in better optimization. The joint training has improved the performance of the ASR module significantly as the additional MT module has provided better (light) supervision in the context of erroneous ASR transcripts. Adding the punctuation information into the input text improves the performance of the MT-model greatly. In line with this observation, use of ASR system generating punctuated text also improves the MT performance significantly in a cascade pipeline. Use of the MT text only data to adapt the ASR decoder to the MT module in the joint-ASR-MT model further improves the performances of these systems. The systems trained in this work are offline models and their performances needs to be studied from the perspective of online or streaming models.

\section{Acknowledgements}
\label{acknow}
The work was supported by Czech National Science Foundation(GACR) project ``NEUREM3'' No. 19-26934X. Part of high-performance computation run on IT4I supercomputer and was supported by the Ministry of Education, Youth and Sports of the Czech Republic through e-INFRA CZ (ID:90140).%, and by European Union’s Horizon 2020 projects No. 864702 - ATCO2. It was also supported by by the Office of the Director of National Intelligence(ODNI), Intelligence Advanced Research Projects Activity (IARPA)MATERIAL program, via Air Force Research Laboratory (AFRL)contract # FA8650-17-C-9118. The views and conclusions contained herein are those of the authors and should not be interpreted as necessarily representing the official policies, either expressed or implied, of ODNI, IARPA, AFRL or the U.S. Government. This work was also supported by The Ministry of Education, Youth and Sports from the Large Infrastructures for Research, Experimental Development and Innovations project ``e-Infrastructure CZ – LM2018140". 

\bibliographystyle{acl_natbib}
\bibliography{IWSLT2021.bib}
\end{document}